\documentclass[11pt]{article}
\usepackage{ranlp2011}
\usepackage{times}
\usepackage{url}
\usepackage{latexsym}
\usepackage[utf8]{inputenc}
\usepackage{textcomp}
\usepackage{amsfonts}
\usepackage{float}
\usepackage{qtree}
\usepackage{graphicx}
\usepackage{listings}
\floatstyle{boxed} 
\restylefloat{figure}

\title{Transition-Based Dependency Parsing With Pluggable Classifiers}

\author{Alex Rudnick\\
  School of Informatics and Computing, Indiana University \\
  Bloomington, Indiana, USA\\
  {\tt alexr@cs.indiana.edu} }

\date{}

\begin{document}
\maketitle

\begin{abstract}
In principle, the design of transition-based dependency parsers makes it
possible to experiment with any general-purpose classifier without other
changes to the parsing algorithm. In practice, however, it often takes
substantial software engineering to bridge between the different representations
used by two software packages. Here we present extensions to MaltParser that
allow the drop-in use of any classifier conforming to the interface of the Weka
machine learning package, a wrapper for the TiMBL memory-based learner to this
interface, and experiments on multilingual dependency parsing with a variety of
classifiers. While earlier work had suggested that memory-based learners might
be a good choice for low-resource parsing scenarios, we cannot support that
hypothesis in this work. We observed that support-vector machines give better
parsing performance than the memory-based learner, regardless of the size of
the training set.
\end{abstract}

\section{Introduction}
Here we present malt-libweka, a library that extends MaltParser to allow users
to experiment with any supervised machine learner compatible with the Weka
machine learning package. This significantly reduces the software engineering
effort required to integrate new classifiers with MaltParser. The Weka
distribution comes with many classifiers, and third-party classifiers may
additionally provide interfaces to Weka. In the cases where they do not, it is
fairly straightforward to implement an appropriate wrapper so that the package
in question can be used with Weka, and so now also with MaltParser. We have
done precisely this with TiMBL, the Tilburg Memory-Based Learner, and the
process is described later in this paper.

With these extensions to MaltParser, we carried out experiments in multilingual
dependency parsing with a variety of classifiers, following the CoNLL-X shared
task \cite{buchholz-marsi:2006:CoNLL-X}. We also generated learning curves for
each classifier, to see how the different algorithms would perform with varying
training data sizes. We had considered the hypothesis, as suggested by earlier
work using several general-purpose classifiers for the same NLP task
\cite{banko-brill:2001:ACL}, that a memory-based learner would provide better
parsing accuracy than MaltParser's default SVM and linear classifiers for small
training sets, but our experiments with the default TiMBL settings do not
support that hypothesis. Instead, we found that, absent any particular
parameter tuning, SVMs gave us the best parsing accuracy for all of our
experimental settings, for each of the four languages in our experiments.

\section{Transition-Based Dependency Parsing}
Transition-based dependency parsers such as MaltParser
\cite{Nivre06maltparser:a} are popular for a number of attractive features.
First, in their deterministic variety, they operate in linear time in the
length of the input sentence, so are comparatively fast when compared with
graph-based or chart-parsing methods, which operate in polynomial time
\cite{DBLP:series/synthesis/2009Kubler}.  Secondly,
transition-based methods give state-of-the-art parsing accuracy in many
settings; in the recent CoNLL shared tasks on multi-lingual dependency parsing,
many of the top-ranking systems were based on transition-based algorithms, and
often used MaltParser specifically
\cite{buchholz-marsi:2006:CoNLL-X,nivre-EtAl:2007:EMNLP-CoNLL2007}.  Of
additional interest is that transition-based parsing algorithms
have an isolated classification task and can make use of 
general-purpose machine learners to address it.
Thus the user of the parser may experiment with different classification
algorithms or parameters while keeping the rest of the parsing system fixed.

Deterministic transition-based dependency parsers come in several varieties,
but in general, they make a single pass over an input sentence, token by token,
and build a dependency structure as the result of a bounded-length series of
decisions. At each point in the processing of a sentence, the parser is said to
be in a given \emph{configuration}, and it must choose which possible
\emph{transition} to make in order to proceed to the next configuration:
eventually, the parser makes its way from the initial configuration to a final
one, in which all of the words in the sentence have been processed. This
parsing approach is analogous to the shift-reduce parsing that one might
use in a constituency parsing setting \cite{DBLP:series/synthesis/2009Kubler}.

Typically, in the initial configuration, an input sentence has been loaded into
a buffer $B$, and there is an empty stack $S$, which will have tokens pushed on
to it and popped off in the subsequent transitions.  Along the way, dependency
arcs are formed between words on the front of the buffer and the top of
the stack, and these are added to $A$, the set of current arcs, which, in
the final configuration, constitutes the dependency parse of the sentence.

There are several possible ``transition systems" that can be used with
transition-based dependency parsing, each of which provides a different set of
transition operations for proceeding through the configurations in the
derivation of a particular parse. Many, but not all, transition systems derive
only \emph{projective} dependency trees, which is to say that for any directed
arc $(w_i, r, w_j)$ (an arc from word $w_i$ to word $w_j$ with dependency
relation $r$), all of the words between $w_i$ and $w_j$ are also either
dependents of $w_i$ or transitively dependent on it. Thus for projective trees,
dependency relations describe contiguous regions where all of the words share
the same head, not unlike the constituents that one might see in a constituency
parsing task.

A transition system for projective dependency trees should be both sound and
complete with respect to the set of projective dependency trees; this is to say
that every output that can be produced by the transition system is in fact a
valid projective dependency tree (\emph{soundness}), and that every projective
dependency tree can be produced by some sequence of transitions from the
transition system (\emph{completeness}). The soundness and completeness proofs
for several transition systems are provided in
\cite{Nivre:2008:ADI:1479202.1479205}.  A baseline transition system with only
three operations (\emph{left-arc}, \emph{right-arc}, and \emph{shift}) is
described in \cite{DBLP:series/synthesis/2009Kubler}; these three transitions
are sufficient to produce any projective dependency tree. The intuition for the
completeness proof is also provided in the book: an algorithm is provided to
map from any projective dependency tree to a sequence of these three
transitions, and therefore a transition sequence exists such that any
particular projective tree could be produced by this transition system.

We have yet to describe the process of how the parsing algorithm decides which
transition to take, out of all possible transitions from the current
configuration. Given an oracle, a parser could make optimal decisions about how
to best proceed to the correct parse; in practice, supervised machine learning
techniques are used to simulate an oracle. The parser has a classifier that has
been trained to predict, for a given configuration, what the best available
transition is. The training data for these classifiers is produced from a
dependency treebank using an algorithm like the one mentioned previously,
mapping from parses of sentences to sequences of (configuration, transition)
pairs. Features are then extracted from the configurations, and the classifier
is trained to predict the transitions, given the features. Commonly used
features include the forms, part-of-speech tags, and dependency relations
associated with the top word of the stack or next upcoming word from the
buffer, though other variations are possible.

While the techniques described so far only produce projective trees, it is
often useful, in describing the syntax of natural languages, to allow
non-projective dependency structures with crossing arcs. Many of the
non-projective structures are familiar from the constituency-parsing world as
those that cause difficulties for context-free grammars. For example, in
English, topicalization or wh-pronouns (in the case of questions) often make
the object of a verb appear outside of a contiguous range with the rest of the
dependents of the verb. Non-projective structures are also common in languages
with more free word order.

There are at least three different ways to produce non-projective dependency
trees with a transition-based parser. One could use a different transition
system with extra operations that produce non-projective trees by moving words
from the stack back on to the buffer, as described in
\cite{DBLP:series/synthesis/2009Kubler}, or one could use a modified parsing
algorithm like that of Covington, which makes use of more than one stack
\cite{Covington01afundamental}. Alternatively, one could use a
``pseudo-projective" approach, where the non-projective structures are
converted to projective ones and annotated in the dependency labels during a
preprocessing step. Then at parse time, the classifier will hopefully predict
the enriched labels when generating projective trees; these labels include
enough information to reconstruct the non-projective trees. This approach
is very effective in practice, and was used for many of the winning CoNLL-X
shared task entries \cite{buchholz-marsi:2006:CoNLL-X}.

In this work, we are concerned with Nivre's Arc-Eager transition system,
initially described in \cite{Nivre03anefficient}, which has the operations
\emph{shift}, \emph{left-arc$_r$}, \emph{right-arc$_r$}, and \emph{reduce}. The
arc-creating operations are parameterized by some dependency relation $r$ from
the set of possible dependency relations $R$, which varies according to the
task or treebank in question. The Arc-Eager transition system, without
modifications, produces only projective dependency trees, but can be used with
pseudo-projective parsing.  Arc-Eager modifies earlier systems that did not
have a separate \emph{reduce} operation, and would eliminate words from the
buffer immediately upon attaching them to their heads, if they appeared to the
right of the head. The Arc-Eager system adds the the \emph{reduce} operation
and thus permits transition sequences in which appropriate arcs can be created
eagerly, with the dependent word being used in subsequent arcs as well, since
its \emph{right-arc} operation does not eliminate the dependent word.

\section{MaltParser}
MaltParser is a popular package for transition-based dependency parsing,
developed by Hall, Nilsson and Nivre\footnote{\url{http://maltparser.org};
version 1.7.1 was used in this work}. MaltParser comes with implementations of
several (nine, as of the current version) transition systems for dependency
parsing; the default is Nivre's Arc-Eager system. MaltParser also
comes with transition systems that can produce non-projective trees, and pre-
and post-processors for pseudo-projective parsing.

For learning to make transition decisions, MaltParser is packaged with two
classifier libraries, LIBSVM \cite{CC01a} and LIBLINEAR \cite{REF08a}. These
packages provide a variety of classification techniques, including support
vector machines with various kernels, linear support vector machines, and
logistic regression. Each of these classifiers has tunable parameters, such as
the type of kernel used for SVMs, or regularization options for
logistic regression. MaltParser uses SVMs with a polynomial kernel by default;
this was the kernel used by the MaltParser team during both of the CoNLL
multilingual dependency parsing shared tasks. In earlier versions of
MaltParser, memory-based learning with TiMBL was also supported
\cite{nivre-hall-nilsson:2004:CONLL}, although this has been removed in the
post-1.0 versions of the system, which are implemented in Java. Previous to
version 1.0, MaltParser was written in C.

MaltParser seems to have been designed with generality and extensibility in
mind; it has has an internal API for integrating arbitrary classifiers, and
much of the program logic has been pushed into separate XML files and expressed
declaratively. However, large portions of the MaltParser code are specific to
LIBSVM and LIBLINEAR, and no documentation about how to add more classifier
libraries is provided, so researchers who wish to experiment with other
classifiers have a significant software engineering task ahead of them.

\section{Weka}
Weka \cite{DBLP:journals/sigkdd/HallFHPRW09} is a popular machine learning
toolkit for Java, freely available
online\footnote{\url{http://www.cs.waikato.ac.nz/ml/weka/}}. It includes
implementations of a variety of machine learning algorithms, and each algorithm
for a given task -- classification, clustering, etc. -- follows a common
interface.  Weka can be used either as a stand-alone application or as a
library for other JVM programs, and for any given task or data set, Weka makes
it convenient to experiment with different machine learners and parameters for
those learners.  Several third-party machine learning packages also include
wrappers for the Weka interface, allowing them to be plugged in to any
application using the Weka standard. This variety and generality makes it seem
like a natural fit with transition-based dependency parsing; we would like to
make it possible to try any classification algorithm as a component of a
parsing system.

One caveat about machine learners included with Weka is that they are not
necessarily high-performance, particularly when compared with the
implementations of the same algorithms from special-purpose packages
such as LIBSVM and LIBLINEAR; while it was easy, from a performance standpoint,
to use the decision-tree and Naive Bayes classifiers from Weka, we could
not get any parses to succeed using Weka's logistic regression classifier, due
to performance problems that will be discussed later. But Weka contains dozens
of other classifiers, and some of them may be perfectly suitable for parsing
with MaltParser.

\section{The CoNLL-X Shared Task}
In the 2006 CoNLL-X shared task on multilingual dependency parsing
\cite{buchholz-marsi:2006:CoNLL-X}, participants built dependency parsing
systems capable of handling many languages, ideally with the same parsing
algorithm and the same machine learners, although perhaps with different
parameter settings per language. The evaluation was carried out over thirteen
different languages, from a variety of language families, although one
(Bulgarian) was optional. The training data made available to the participants
contained some non-projective structures, as did the gold standard parses for
the testing data, though the systems were not strictly required to produce
non-projective parses.

The CoNLL dependency format \cite{buchholz-marsi:2006:CoNLL-X} has become a
standard for dependency parsers. CoNLL-formatted parse trees separately
describe each token of a sentence, and can include raw and lemmatized versions
of each word, coarse- and fine-grained part of speech tags, additional lexical
features, the head of the token, and token's dependency relation to the head.
The lexical features present vary for each language, but they might include
information like number, gender, and case. In practice, these features do not
seem to be used by working parsers -- MaltParser comes with feature sets that
make use of in parts of speech, dependency relations, and the surface forms of
words. 

The participants presented systems based on a variety of techniques, but the
best systems used either transition-based dependency parsing or graph-based
strategies like those of McDonald's MSTParser
\cite{mcdonald-EtAl:2005:HLTEMNLP}. Among the transition-based systems, the
highest-scoring parsers used pseudo-projective strategies, deterministic
parsing algorithms, and support vector machines with polynomial kernels.

\section{Experiments}
\begin{figure*}
\begin{center}
\includegraphics[width=7.9cm]{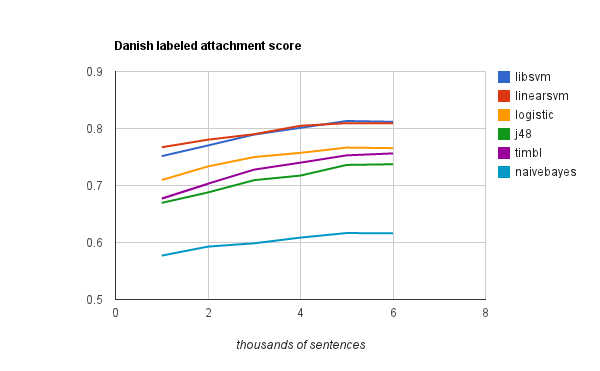} \includegraphics[width=7.9cm]{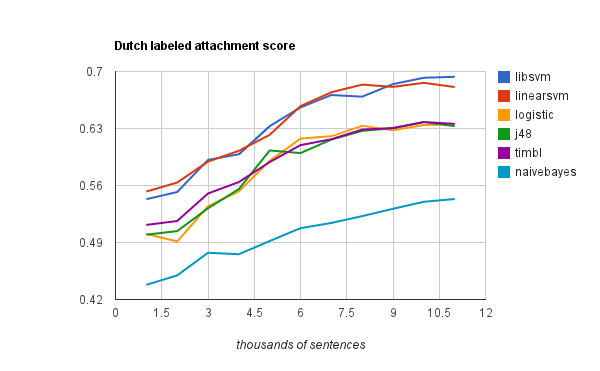} 
\includegraphics[width=7.9cm]{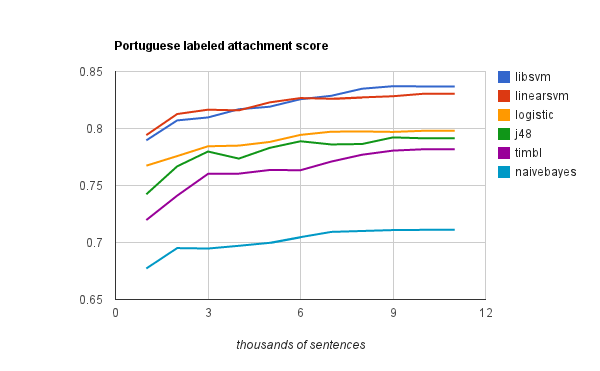} \includegraphics[width=7.9cm]{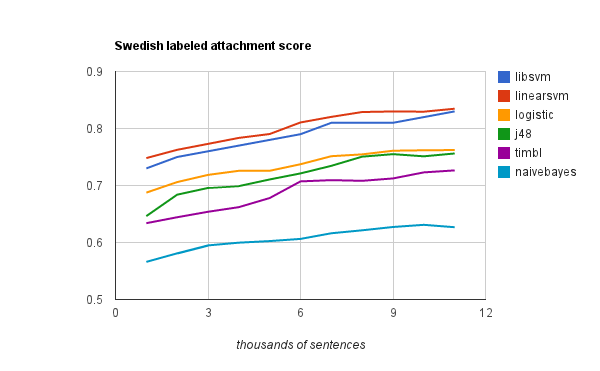} 
\end{center}
\caption{LAS learning curves for the four languages and various classifiers.}
\label{curves}
\end{figure*}

\begin{figure*}
\begin{center}
\includegraphics[width=12cm]{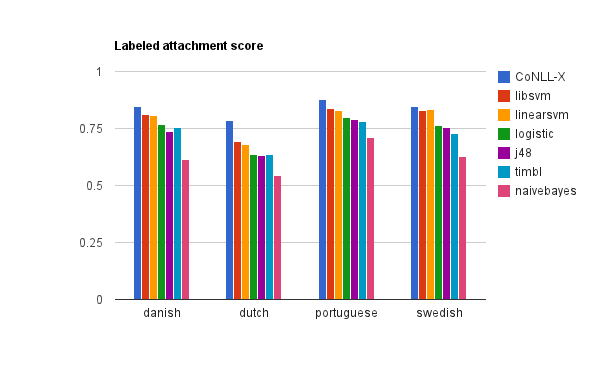}
\end{center}
\caption{Labeled attachment score for the various languages and classifiers, at
the maximum training corpus size, with the winning CoNLL-X scores, for
comparison.}
\label{scores}
\end{figure*}

To evaluate a variety of classifiers, we replicated the CoNLL-X (2006) shared
tasks on multilingual dependency parsing. In the interests of reproducibility
and frugality, here we ran experiments on the languages with freely available
treebanks\footnote{The freely-available data for the CoNLL-X shared task is
online at \url{http://ilk.uvt.nl/conll/free_data.html}}. Out of the thirteen
languages in the evaluation, this leaves Danish, Dutch, Portuguese and Swedish.

We did almost no parameter tuning or feature engineering, save making sure that
the task could be run in 8 gigabytes of RAM. We used a feature set already
present in MaltParser -- the one used for parsing with the Arc-Eager transition
system and LIBSVM -- and default settings for each software package to get an
initial sense for each classifier's behavior. There are almost certainly
classifier settings and feature sets that would provide better parsing
accuracy, but finding those is left to interested parties in the future.

Additionally, we did not use pseudo-projective post-processing, since the goal
of the experiments was simply to compare the available classifiers. The best
CoNLL-X entries performed rather better than the parsers trained in these
experiments, and this is due in part to their handling of non-projective
structures, which is described in \cite{nivre-EtAl:2006:CoNLL-X}.

We also generated learning curves for the parsing task, varying the amount of
training data given to the classifiers in increments of a thousand sentences,
from one thousand sentences, up to eleven thousand, which was roughly the
entire training set for two of the four languages. For Danish, however, the
training set was only 5190 sentences long, so the experiments for that language
cut off at six repetitions, and for Portuguese, the training set contains 9071
sentences, so the tenth and eleventh iterations were the same. In all cases, we
show the labeled attachment score curves in Figure \ref{curves}. The unlabeled
attachment scores varied similarly, and of course were higher; some of these
are given in Figure \ref{numbers}. In Figure \ref{scores} we also show the best
available CoNLL-X labeled attachment scores for comparison.

We report results for six different classifiers, which are listed in Figure
\ref{classifiers}. Three of the classification approaches that we tested,
LIBSVM with a polynomial kernel, LIBLINEAR's linear support vector machines,
and LIBLINEAR's logistic regression (also known as a ``Maximum Entropy"
classifier), are available by default with MaltParser. The remaining three made
use of the malt-libweka interface; J48 decision trees and the Naive Bayes
classifier are familiar algorithms that have implementations in Weka. The TiMBL
memory-based learner also made use of the malt-libweka interface, through a
new wrapper that was implemented for these experiments.

Looking at the results, we see that, in all cases, the SVM classifiers
outperform the other classifiers, typically followed by logistic regression,
decision trees, and TiMBL. Perhaps unsurprisingly, Naive Bayes does not give
good parsing results on these tasks, and gave the worst performance for all
settings; the features in a parsing task are not mutually independent. At
a few points in the learning curves, logistic regression is met or outperformed
by TiMBL or decision trees, but at all of the points along the curves,
the SVM classifiers perform several LAS points better than the next-best
classifier.

We did observe, however, that across all four languages, the linear SVMs
outperformed the polynomial-kernel SVMs when trained with the smallest corpora.
Also, in parsing Swedish, the linear SVMs were consistently slightly better
than the polynomial-kernel ones. The higher performance of the linear SVM on
the smaller data sets could be explained by its higher \emph{bias}, which is to
say that, while it is less able to express complex hypotheses in
higher-dimensional spaces, this makes it less likely to over-fit small training
sets, so the result is not entirely surprising.

Our confidence that the implementation of malt-libweka is basically correct,
and not the source of the lower performance of the other classifiers, stems
from comparing the performance of the J48 and TiMBL classifiers with that of
the linear regression setting for LIBLINEAR; they had very comparable
performance with one of the major modes of operation for LIBLINEAR, in some
cases equalling or outperforming it. It seems that support vector machines are
simply a good choice for parsing tasks.  

\begin{figure*}
\begin{center}

\begin{tabular}{l | l }
  label & description \\
  \hline 
  libsvm     & LIBSVM: polynomial kernel (default for MaltParser) \\
  linearsvm  & LIBLINEAR: linear SVM \\
  logistic   & LIBLINEAR: logistic regression \\
  j48        & malt-libweka with Weka's J48 decision tree classifier \\
  timbl      & malt-libweka with our new Weka wrapper for TiMBL \\
  naivebayes & malt-libweka with Weka's NaiveBayes classifier \\
\end{tabular}

\end{center}
\caption{The six different classifiers used in experiments.}
\label{classifiers}
\end{figure*}

\begin{figure*}
\begin{center}
\begin{tabular}{l| r|r|r|r|r|r|r|r}
  classifier & da (sm) & nl (sm) & pt (sm) & sv (sm)
             & da (lg) & nl (lg) & pt (lg) & sv (lg)  \\
  \hline 
  libsvm     & 75, 81 & 54, 59 & 79, 84 & 73, 81
             & 81, 86 & 69, 72 & 84, 80 & 83, 89 \\
  linearsvm  & 77, 84 & 55, 62 & 79, 86 & 75, 84
             & 81, 86 & 68, 72 & 83, 88 & 83, 89 \\
  logistic   & 71, 79 & 50, 57 & 77, 83 & 69, 78
             & 77, 83 & 64, 68 & 80, 84 & 76, 84 \\
  j48        & 67, 75 & 50, 58 & 74, 81 & 65, 74
             & 74, 82 & 63, 68 & 79, 85 & 76, 84 \\
  timbl      & 68, 76 & 51, 58 & 72, 80 & 63, 75
             & 76, 83 & 64, 68 & 78, 85 & 73, 82 \\
  naivebayes & 58, 66 & 44, 52 & 68, 75 & 57, 64
             & 62, 69 & 54, 58 & 71, 77 & 63, 69 \\
\end{tabular}

\end{center}
\caption{Scores for different classifiers (rounded, as a percentage), on each
of the four languages, Danish, Dutch, Portuguese, and Swedish. The designation
(sm) is for the smallest training set that was tried with the given language,
and (lg) indicates the largest. In all cells of the table, the first number is
the LAS, and the second is the UAS.}
\label{numbers}
\end{figure*}

\section{Software}
One contribution of this work is a reusable package, malt-libweka, which is
freely available
online\footnote{\url{http://github.com/alexrudnick/malt-libweka}}.
malt-libweka itself is a library that works with MaltParser; its repository
includes scripts that can be used to reproduce the results in this paper, or
could be easily modified to do further parsing experiments on treebanks in the
CoNLL format.

While MaltParser is open-source and designed for extensibility, the development
of malt-libweka took non-trivial software engineering effort, largely
due to a lack of documentation of the internals of MaltParser, which was
perhaps not implemented with the convenience of third-party developers in mind.
While MaltParser apparently has a plugin system, both mentioned in the
online documentation and present in the source code, it was non-obvious 
how to use it, and we could not find examples of it being used in
practice. Whether or not the plugin system is in a usable state,
MaltParser's source tree definitely contains substantial amounts of ``dead
code" with misleading names. Particularly, while the ``LibSvm" and
``LibLinear" classes are instrumental in MaltParser's interfaces to the
corresponding machine learning packages, MaltParser also contains the classes
``Libsvm" and ``Liblinear" -- note the capitalization differences. The latter
two seem to be entirely vestigial, and not called in the current version.

Hopefully the use of malt-libweka will save future developers from having to
delve too much into the source of MaltParser when they would like to experiment
with MaltParser and different classifiers. With malt-libweka, users need only
adapt their machine learners to the interface used in Weka; a straightforward
example of this is provided in the \texttt{maltparser.TimblClassifier} class.
For this use case, the only required methods in the interface are
\texttt{buildClassifier} and \texttt{classifyInstance}, which, respectively,
train a classifier given a set of training instances and return a predicted
class for a given instance. The existing MaltParser code, coupled with
malt-libweka, handle the rest of the process, including extracting the relevant
features from parse configurations and then making those features available to
the machine learner, both at training and parsing time.

\subsection{Implementing the TiMBL-Weka Interface}
TiMBL, described in detail in \cite{daelemans2010timbl}, is a package for
memory-based learning, and is freely available
online\footnote{\url{http://ilk.uvt.nl/timbl/}}.  As a lazy learner, TiMBL's
training process consists of storing all of the examples that it is
given for use at classification time, where the values of the features of these
examples could be treated in a symbolic, nominal way, or as numbers over which
there is an ordering or a distance.  At an implementation level, the TiMBL
classifier can be run as a server with the timblserver package, making it
possible to interface with TiMBL from a program written in any language that
has a networking library.

In implementing the Weka wrapper for TiMBL, we had to implement the code for
training time, and for classification time. For the code for the training
procedure, we serialize all of the training instances to a file readable by the
TiMBL software, which uses, roughly, a CSV format. Then, when it is time to
parse, we need to be able to classify new instances, so the Java program opens
a connection to the timblserver -- which must be started by some external
program, in this case the scripts that manage the parsing experiments -- and
serializes a new instance then sends it across the network. Then TiMBL sends
back its classification result, and we must, on the Java side, reinterpret the
result as a number for MaltParser's consumption. The implementation of the Weka
wrapper for TiMBL took roughly 100 lines of Java, most of which manages the
network connection.

\section{Discussion}
An issue that we encountered during development was that we had to maintain the
meaning of the representations used internally by MaltParser, when they were
passed to Weka classifiers. At an early stage of processing, MaltParser builds
several vocabularies, mapping from tokens and the provided lexical features
(POS tags, etc) to integers, so that the system need not pass around large
strings. So sensibly, both at training time and at parsing time, the
classifiers called by MaltParser are presented with integers rather than
strings. These should not be treated as anything other than unique identifiers,
but it would be fairly easy for a programmer to allow a classifier trained on
these numbers to interpret them as ordinal numbers or take distances over them.
But during early development of the system, we made exactly this mistake; we
discovered the problem when inspecting the decision trees learned by Weka's J48
classifier, which was making comparisons with a less-than operator. J48 is a
Java reimplementation of the C4.5 algorithm \cite{Quinlan:1993:CPM:152181},
which will try to do comparisons over ordinal numbers given the opportunity.
With this in mind, we made Weka interpret the features passed to it as nominal
features -- although they are still represented as numbers -- which prevents
order-based comparisons.

However, the logistic regression algorithm, in a mathematical sense, is defined
in terms of distances over numbers. If the Weka implementation is given nominal
attributes, it will binarize them into a larger number of binary attributes. In
this process, an attribute that has $n$ possible values is transformed into $n$
different binary attributes. So for many features passed to the learner during
the parsing task, there are many thousands of possible values.  If we consider
the feature ``which word is on the top of the stack", any word in the
vocabulary could appear.

During development, we ran across a few surprising performance problems. The
binarization code in Weka is much less efficient than it could be, and while
trying to parse some of the smaller datasets, the system would run out of
memory during binarization, even when given 8 gigabytes of RAM. This seemed
surmountable, so we implemented a more efficient version of feature
binarization (\texttt{maltlibweka.FastBinarizer}), in hopes that this would let
us experiment with Weka's logistic regression.  But the training times for
Weka's \texttt{Logistic} class ended up being unbearably long and prohibitively
memory-intensive when given large numbers of binary features, so we also tried
a few approaches for feature selection, though were not successful in this
regard. In the end, we gave up on Weka's logistic regression implementation,
although we had hoped to compare it to the one in LIBLINEAR.

So while any given classifier may not perform well in terms of parsing
accuracy, or even computational efficiency -- as we have seen in the course of
this work -- malt-libweka makes it straightforward to try new classifiers and
new parameters for those classifiers on parsing tasks. And to adapt a new
classifier to work with Weka and thus malt-libweka, the programmer need only
provide a method that trains the classifier given a set of training instances
and another that classifies a given instance after training, barring mishaps
with the classifier not scaling well to the parsing task.

\section{Conclusions and Future Work}
We have introduced extensions to MaltParser that enable experimentation with
different classifiers for transition-based dependency parsing, making such
experiments straightforward in practice, when previously they were only
straightforward in theory. We have also presented experiments with six
classifiers for a standard multilingual dependency parsing task, including
varying the size of the training set. We were not able to support the
hypothesis that memory-based learners provide better parsing accuracy than
support-vector machines in low-resource settings; in fact, for settings with
small training sets as well as those with comparatively large ones, support
vector machines continue to perform the best out of the approaches considered.
Our results also suggest that for small training sets, linear support vector
machines are a good choice.

There may well be classifiers, or parameter settings for the algorithms, that
learn better parsers for training sets of these sizes for these languages.
There may also be better feature sets, perhaps making use of agreement
information for morphologically rich languages, and those with more free word
order. Finding out which parameters and which settings is, however, left to
future work. Hopefully malt-libweka will make these experiments easy to carry
out.

\bibliographystyle{acl}
\bibliography{parsing.bib}{}
\end{document}